\documentclass[letterpaper]{article} 

\usepackage{aaai2026}  
\usepackage{times}  
\usepackage{helvet}  
\usepackage{courier}  
\usepackage[hyphens]{url}  
\usepackage{graphicx} 
\usepackage{natbib}  
\usepackage{caption} 
\usepackage{tikz}
\usepackage{graphicx}
\usetikzlibrary{positioning, shapes.misc, shadows}
\usepackage{float} 

\usepackage{amsmath}        
\usepackage{amssymb}        
\usepackage{amsfonts}
\usepackage{csquotes}
\usepackage{algorithm}      
\usepackage{algpseudocode}    

\usepackage[colorinlistoftodos]{todonotes}

\usepackage{newfloat}
\usepackage{listings}
\usepackage{booktabs}
\usepackage{array}
\usepackage{tcolorbox}
\usepackage{dirtytalk}
\usepackage{tabularx}
\usepackage{makecell}

\newcommand{\ekb}{\emph{EKB}}

\DeclareCaptionStyle{ruled}{labelfont=normalfont,labelsep=colon,strut=off} 
\floatstyle{ruled}
\newfloat{listing}{tb}{lst}{}
\floatname{listing}{Listing}
\title{Beyond the Pixels: VLM-based Evaluation of Identity Preservation in Reference-Guided Synthesis}

\author{
Aditi Singhania\textsuperscript{\rm 1},
Krutik Malani\textsuperscript{\rm 1},
Riddhi Dhawan\textsuperscript{\rm 1},
Arushi Jain\textsuperscript{\rm 1},
Garv Tandon\textsuperscript{\rm 1},
Nippun Sharma\textsuperscript{\rm 1},
Souymodip Chakraborty\textsuperscript{\rm 1},
Vineet Batra\textsuperscript{\rm 1},
Ankit Phogat\textsuperscript{\rm 1}
}
\affiliations{
\textsuperscript{\rm 1} Adobe  \\
}

\nocopyright
\begin{document}
\maketitle

\begin{abstract}
Evaluating identity preservation in generative models remains a critical yet unresolved challenge. Existing metrics rely on global embeddings or coarse VLM prompting, failing to capture fine-grained identity changes and providing limited diagnostic insight. We introduce \textbf{Beyond the Pixels}, a hierarchical evaluation framework that decomposes identity assessment into feature-level transformations. Our approach guides VLMs through structured reasoning by (1) hierarchically decomposing subjects into (type, style) → attribute → feature decision tree, (2) prompting for concrete transformations rather than abstract similarity scores. This decomposition grounds VLM analysis in verifiable visual evidence, reducing hallucinations and improving consistency.  We validate our framework across four state-of-the-art generative models, demonstrating strong alignment with human judgments in measuring identity consistency. Additionally, we introduce a new benchmark specifically designed to stress-test generative models. It comprises 1,078  image-prompt pairs spanning diverse subject types---including underrepresented categories such as anthropomorphic and animated characters---and captures an average of six to seven transformation axes per prompt.
\end{abstract}

\begin{figure*}[t]
    \centering
    \includegraphics[width=\textwidth]{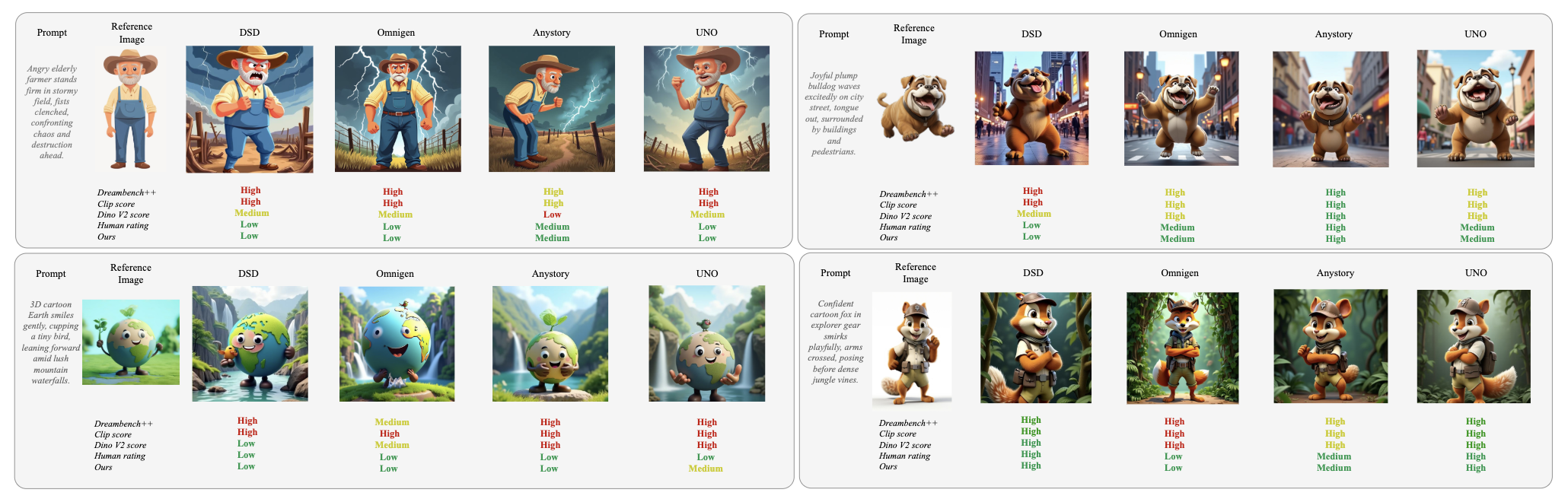}
    \caption{Comparison of different generative models (DSD, Omnigen, Anystory, UNO) illustrating gaps between automated metrics and human judgment in character identity.}
    \label{fig:compare}
\end{figure*}

\section{Introduction}
\textbf{Character consistency} is essential across creative workflows, from storytelling and animation to game design and branding. \textit{Imagine your favorite superhero unexpectedly appearing with altered facial features or costume colors across scenes without narrative requirement}—such inconsistencies disrupt narrative coherence and diminish audience engagement. Recent diffusion-based generative models have significantly advanced personalized character synthesis, transitioning from methods reliant on extensive fine-tuning, specialized encoders, or multiple exemplars to enabling synthesis from merely a single reference image~\cite{wu2025less,cai2024diffusion,tan2024ominicontrol,xiao2025dreamo,tao2025instantcharacter}. 
Moreover, modern diffusion methods can place these characters into complex scenes and novel poses, significantly broadening their applicability across creative media ~\cite{wu2025less,wang2024msdiffusion,xiao2025dreamo}. Despite these advancements, a fundamental question remains: 
\emph{How can we quantitatively measure the quality of generative models on the task of identity preserving generation?}

\section{Related Work}
The rapid advancement of generative AI has necessitated corresponding progress in evaluation methodologies, particularly for identity-preserving image generation where subtle visual changes determine success. We review existing approaches and their limitations.

\paragraph{Global Embedding-Based Similarity.}
Traditional similarity metrics rely on global feature representations that inherently lose fine-grained identity information. CLIP~\cite{radford2021clip} and DINOv2~\cite{oquab2023dinov2} compress images into single embedding vectors, prioritizing coarse semantic alignment over localized feature preservation. This compression rewards shape overlap while overlooking critical discrepancies in facial landmarks, textures, or regional attributes. Recent diagnostic studies~\cite{abbasi2025} demonstrate this vulnerability to attribute-binding failures, where models cannot correctly associate local properties with their spatial regions (e.g., conflating \say{purple sphere} with \say{yellow sphere}).

\paragraph{Single-Image VLM Evaluation.}
Vision-Language Models have emerged as evaluation tools for text-to-image generation, with benchmarks like TIFA~\cite{hu2023tifa}, GenAI-Bench~\cite{li2024genai}, and T2I-FineEval~\cite{hosseini2025} assessing prompt adherence through structured question-answering. However, these frameworks operate on single image-prompt pairs and lack mechanisms for cross-image reasoning. Identity preservation requires joint analysis across reference and generated images; a fundamentally different task that requires correspondence reasoning, transformation understanding, and robustness to stylistic variations that single-image evaluators cannot address.

\paragraph{Holistic Multi-Image Assessment.}
Recent work extends VLM evaluation to multi-image scenarios. DreamBench++~\cite{peng2024dreambenchpp} designs prompting for identity consistency evaluation that aligns with human judgments but employs coarse-grained prompting that asks models to assess identity preservation holistically. Such broad queries (\say{Does this image preserve the subject's identity?}) elicit superficial responses that overlook fine-grained inconsistencies. When multiple visual factors change simultaneously—pose, expression, lighting—VLMs resort to generic assessments rather than systematic attribute-level analysis. Figure~\ref{fig:compare} depicts few examples of sub-optimal evaluation by existing methods.

\paragraph{The Need for Structured Evaluation.}
These limitations stem from a fundamental mismatch between the task complexity and evaluation granularity. 
Our key insight is that VLMs produce more accurate outputs when constrained to evaluate narrow, specific features rather than holistic identity. Furthermore, by prompting VLMs to identify transformations rather than direct similarity scores, we force deeper visual reasoning
that distinguishes between different types of change.

\paragraph{Contributions.}
We make two primary contributions:\newline
\textbf{Structured VLM Evaluation Framework:} We introduce \emph{Comprehensive Human-
Aligned Reference Image Similarity} (CHARIS), a hierarchical framework that decomposes identity assessment into feature-level transformations rather than holistic similarity scores, reducing VLM hallucination and improving evaluation reliability.\newline
\textbf{Comprehensive Benchmark:} We present a benchmark of 1,078 prompts spanning 154 subjects with balanced representation across subject types (human, animal, anthropomorphic, animated) and artistic styles (photo realistic, cartoon, vector), each incorporating 6-7 simultaneous transformations, significantly exceeding the complexity of existing benchmarks.


\section{Problem Definition}
Given a reference image $I_1 \in \mathbb{R}^{H \times W \times 3}$ that defines the identity of subject  and a textual prompt $p$ that specifies target context, a generative model $G$ synthesizes:
\begin{equation}
I_2 = G(I_1, p)
\end{equation}
The task of measuring \emph{core visual identity preservation} between $I_1$ and $I_2$ is inherently challenging due to multiple factors. Prompts may induce \emph{concurrent transformations} in pose, viewpoint, expression, and scene composition, while evaluation must span \emph{diverse styles} from photorealistic to cartoon renderings. Generation artifacts such as \emph{missing accessories, occlusions, or altered details} create ambiguity in identity assessment. 


%
These inherent challenges are amplified when using Vision-Language Models for direct similarity assessment. When prompted with high-level questions like \enquote{rate identity preservation from 1-10}, VLMs exhibit following systematic failures:
\begin{itemize}
    \item \textbf{Cognitive overload:} The simultaneous evaluation of multiple transformation factors leads VLMs to produce shallow, impression-based responses rather than systematic analysis
    \item \textbf{Ambiguous grounding:} Without constraining VLMs to specific features, they default to global pattern matching, missing critical fine-grained inconsistencies
    \item \textbf{Shortcut reasoning:} Direct scoring allows VLMs to bypass detailed visual inspection, generating plausible but generic ratings that fail to capture nuanced identity changes
\end{itemize}
%
This combination of inherent complexity and VLM limitations necessitates a structured approach that decomposes identity evaluation into feature-specific assessments and replaces abstract similarity scoring with concrete transformation identification—forcing VLMs to ground their analysis in verifiable visual evidence.

\section{Hierarchical Decomposition and Transformation-Based Reasoning}
We address these limitations through two key innovations:

\paragraph{Hierarchical Feature Decomposition.}
Rather than asking VLMs to evaluate identity holistically, we decompose subjects into a hierarchy: (type, style) → attributes → features. This decomposition serves to:
\begin{itemize}
    \item \textbf{Ground VLM responses:} By concentrating on specific features—\enquote{eye shape} instead of the broader \enquote{face}—we anchor the VLM’s judgment in verifiable visual evidence. This narrower focus curtails the model’s tendency to \emph{hallucinate} plausible-but-incorrect global assessments.
    \item \textbf{Enable systematic coverage:} The hierarchical structure ensures that every identity-relevant feature is evaluated.
\end{itemize}

\paragraph{Transformation-Based Evaluation.}
Instead of asking \enquote{how similar are these features?}, we prompt VLMs to identify \enquote{what transformations occurred between source and target?} This approach:
\begin{itemize}
    \item \textbf{Forces deeper reasoning:} Identifying specific transformations demands careful visual comparison rather than surface-level matching, which blocks shortcut reasoning: VLMs cannot fallback on generic similarity templates and must instead articulate the concrete visual changes.
    \item \textbf{Enables categorization:} Transformations can be classified (pose-induced, style-induced, intrinsic), providing interpretable output.
    \item \textbf{Improves consistency:} Concrete transformation descriptions are more stable across prompting variations than abstract scores
\end{itemize}

\paragraph{External Knowledge Base Integration.}
Our EKB provides structured priors about valid features and transformations for each subject type and style, further constraining VLM outputs to meaningful visual attributes. This prevents the model from hallucinating irrelevant features or transformations.

Through this structured approach, we transform the ill-posed problem of identity evaluation into a series of well-defined, narrow visual reasoning tasks that VLMs can reliably perform. The aggregation of these localized assessments yields robust identity preservation labels that align with human judgment while providing interpretable diagnostic information.

\section{Definitions}
Let $I : \Omega \subset \mathbb{R}^{2} \!\to\! \mathbb{R}^{3}$ denote a digital image on a 
discrete pixel domain~$\Omega$. We are interested in images that
have subjects. Let $\mathcal{S}_I$ the set of subjects of an image $I$, where each subject 
$s\in\mathcal{S}_I$ is subset of pixels of the domain of $I$. 
Unless stated otherwise, we will assume that $\mathcal{S}$ is singleton, that is, the image 
contains exactly one subject. Given two images $I_1$ and $I_2$, with subjects $s_1$ and $s_2$, 
respectively, our objective is to develop a method to measure the degree of \emph{identity} preservation. 
\[
\varphi(I_1, I_2) \in \mathcal{C}
\] $\mathcal{C} = \{\texttt{exact},\texttt{near\_exact}, \texttt{partial}, \texttt{mismatch}\}$ is the category 
defining similarity between the subjects of the two images.

We systematically decomposed the notion of \emph{identity}. 
First, we define the \emph{type} function  
$\tau: \Omega \to \mathfrak{T}$ that defines the type of the subject and \emph{style} function 
$\kappa:\Omega \to \mathfrak{K}$ that defines the style of the image. 
The type set $\mathfrak T$ consists of \emph{humanoid}, \emph{animals}, \emph{anthropomorphic} and \emph{animated inanimate}. The style set $\mathfrak K$ consists of  \emph{photo realistic}, \emph{vector} and \emph{cartoon}. 

For each subject type and style we have attributes that \emph{identify} the subject. For example, an \emph{animal} 
type with \emph{cartoon} style have \emph{species specific element} and the \emph{Cartoon-Style} as attributes (see Table~\ref{tab:cartoon_animal_ekb_full}). Each attribute then dictates the presence of specific features. 

We will consider the following \emph{Transformations} $X$ defined at the feature level:  (1) \emph{pose variations} (body orientation, limb positioning), (2) \emph{facial expressions} (emotional states, mouth/eye configurations), (3) \emph{viewpoint changes} (frontal to profile, viewing angles), (4) \emph{occlusion patterns} (partial visibility, object obstruction), (5) \emph{lighting conditions} (directional changes, intensity variations), (6) \emph{background contexts} (environmental settings, compositional elements), and (7) \emph{stylistic interpretations} (rendering techniques, artistic mediums).
These transformations 
specify the necessary changes required to modify the appearance of a feature belonging to subject 
$s_1$ in image $I_1$ to match the appearance of the corresponding feature in subject $s_2$ from 
image $I_2$. 
%
The decomposition methodology and transformation definitions are image-agnostic and together constitute the \emph{External Knowledge Base} (EKB). Refer to the supplementary for complete specifications.

\begin{figure}
    \centering
    \includegraphics[width=0.9\linewidth]{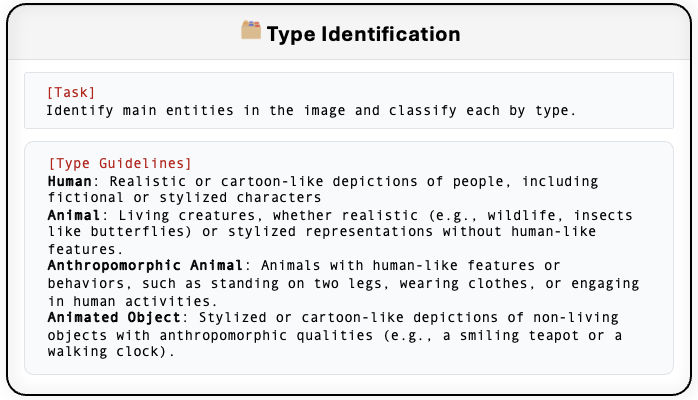}
    \caption{Prompt for extracting Type}
    \label{fig:type_prompt}
\end{figure}

\section{Method}
Our approach for evaluating subject identity preservation across images employs a \emph{chain of thoughts} prompting strategy. The decomposition of a given subject using the External Knowledge Base (\ekb) is accomplished through iterative prompting of a Vision Language Model (VLM) $\theta$. 
Given an image $I$ containing a subject $s$, our method systematically extracts the visible features of the subject through the following sequential prompting process:

\begin{figure}
    \centering
    \includegraphics[width=0.9\linewidth]{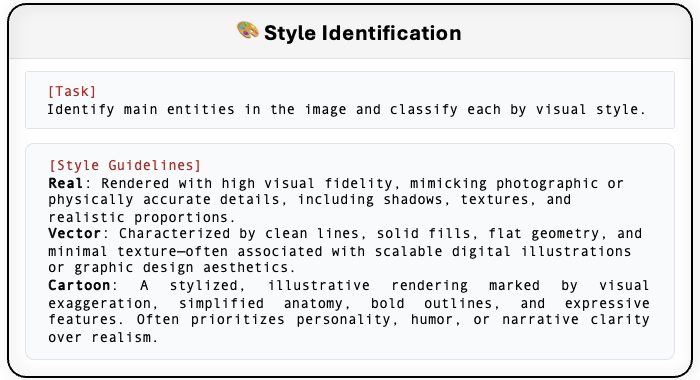}
    \caption{Prompt for extracting Style}
    \label{fig:style_prompt}
\end{figure}

\begin{figure}
    \centering
    \includegraphics[width=0.9\linewidth]{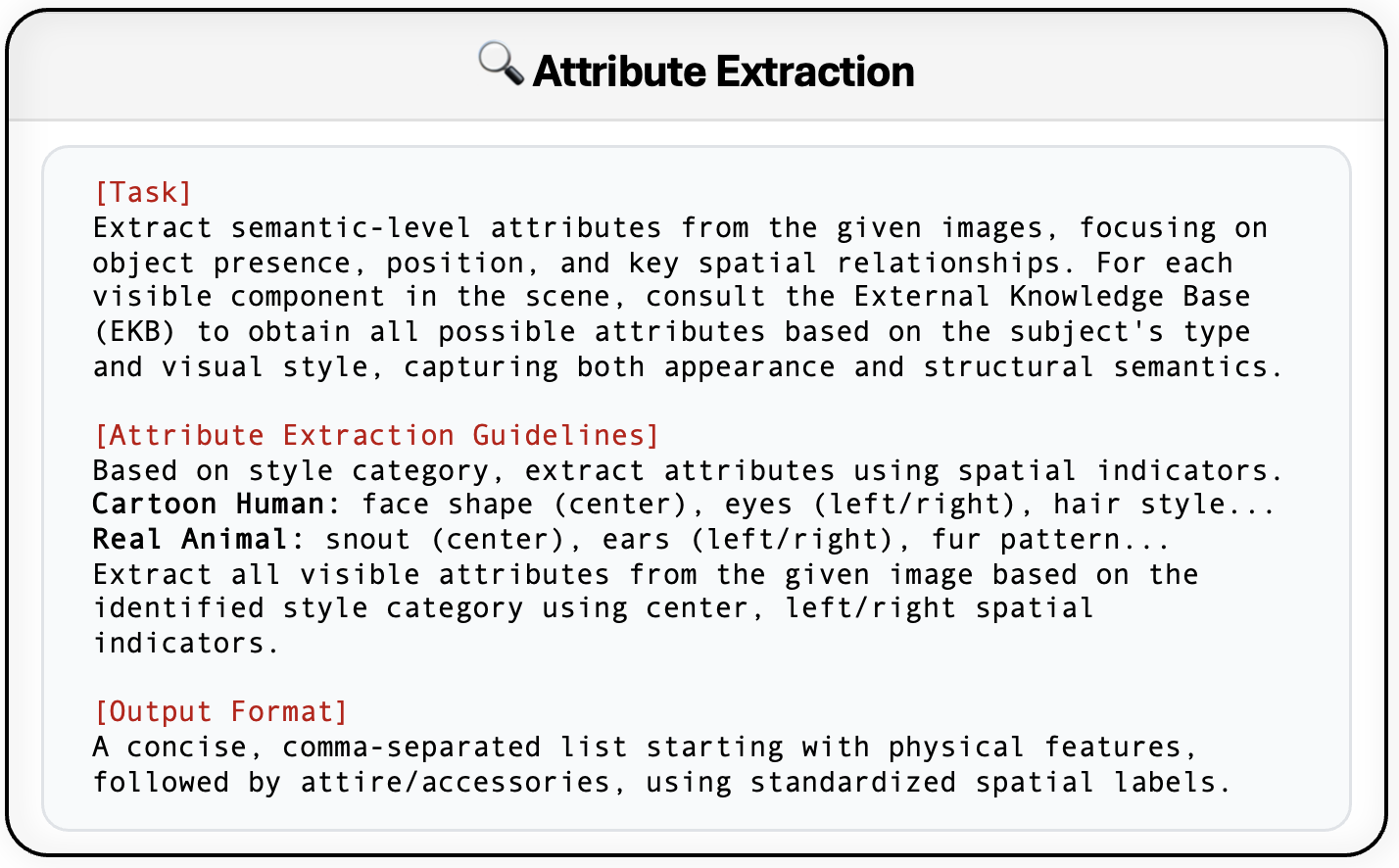}
    \caption{Given a specific style and type we consult \ekb~to create a prompt to extract visible attributes.}
    \label{fig:attr_promp}
\end{figure}

\begin{figure}
    \centering
    \includegraphics[width=0.9\linewidth]{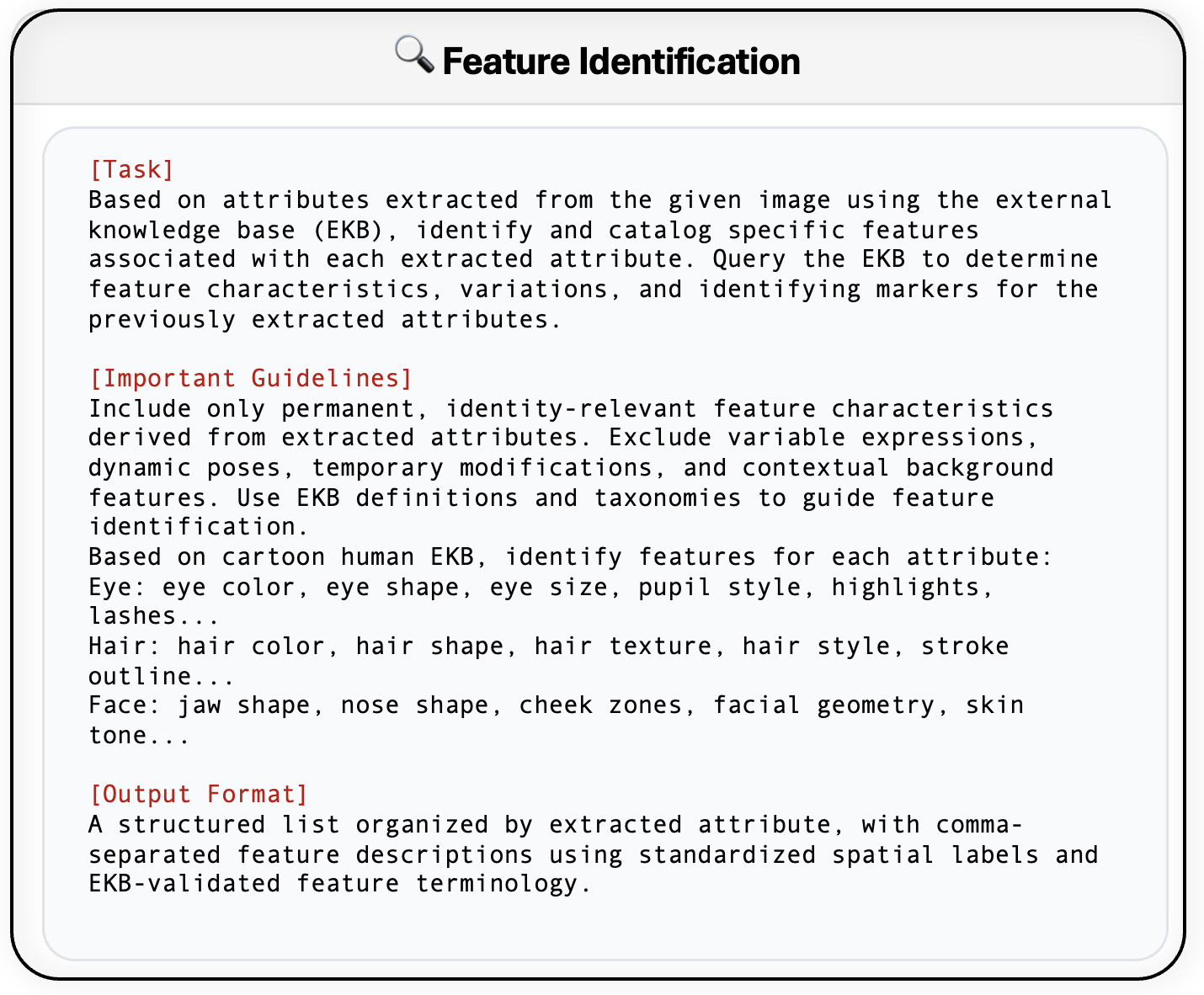}
    \caption{Given visible attributes we consult \ekb~to create a prompt to extract visible features.}
    \label{fig:feature_promp}
\end{figure}

\begin{enumerate}
    \item \textbf{Type and Style Identification:} Starting with image $I$, we determine the subject type $\tau_s := \tau(I) \in \mathfrak{T}$ and style $\kappa_s := \kappa(I) \in \mathfrak{K}$  by prompting $\theta(I, p_1)$ and $\theta(I, p_2)$ respectively. The specific prompts $p_1$ and $p_2$ are detailed in Figures~\ref{fig:type_prompt} and~\ref{fig:style_prompt}. These prompts incorporate information about the type set $\mathfrak{T}$ and style set $\mathfrak{K}$ from the \ekb.
    
    \item \textbf{Attribute Detection:} For the identified type $\tau_s$ and style $\kappa_s$, we retrieve the complete set of possible attributes $A$ from the \ekb.
    We then obtain the \emph{visible} attributes $A_s$ present in the image by prompting the VLM: $C_s := \theta(I, p)$. An example prompt construction is shown in Figure~\ref{fig:attr_promp}.
    
    
    \item \textbf{Feature Identification:} Finally, we obtain the candidate feature set $F$ associated with the visible attributes $A_s$ from the \ekb, and identify the actual visible features via $F_s := \theta(I, p)$. The prompt is constructed using the attributes $A_s$, as shown in Figure~\ref{fig:feature_promp}.
\end{enumerate}
\noindent
\textbf{Cross-Image Identity Evaluation:} Given two images $I_1$ and $I_2$ with subjects $s_1$ and $s_2$ respectively, where $I_1$ serves as the reference (source) image and $I_2$ as the generated (target) image, we define the complete decomposition of image $I_i$ as:
$$H_i := (\tau_{s_i}, \kappa_{s_i}) \to  A_{s_i} \to F_{s_i}$$
where $i \in \{1, 2\}$.

\begin{figure}
    \centering
    \includegraphics[width=0.9\linewidth]{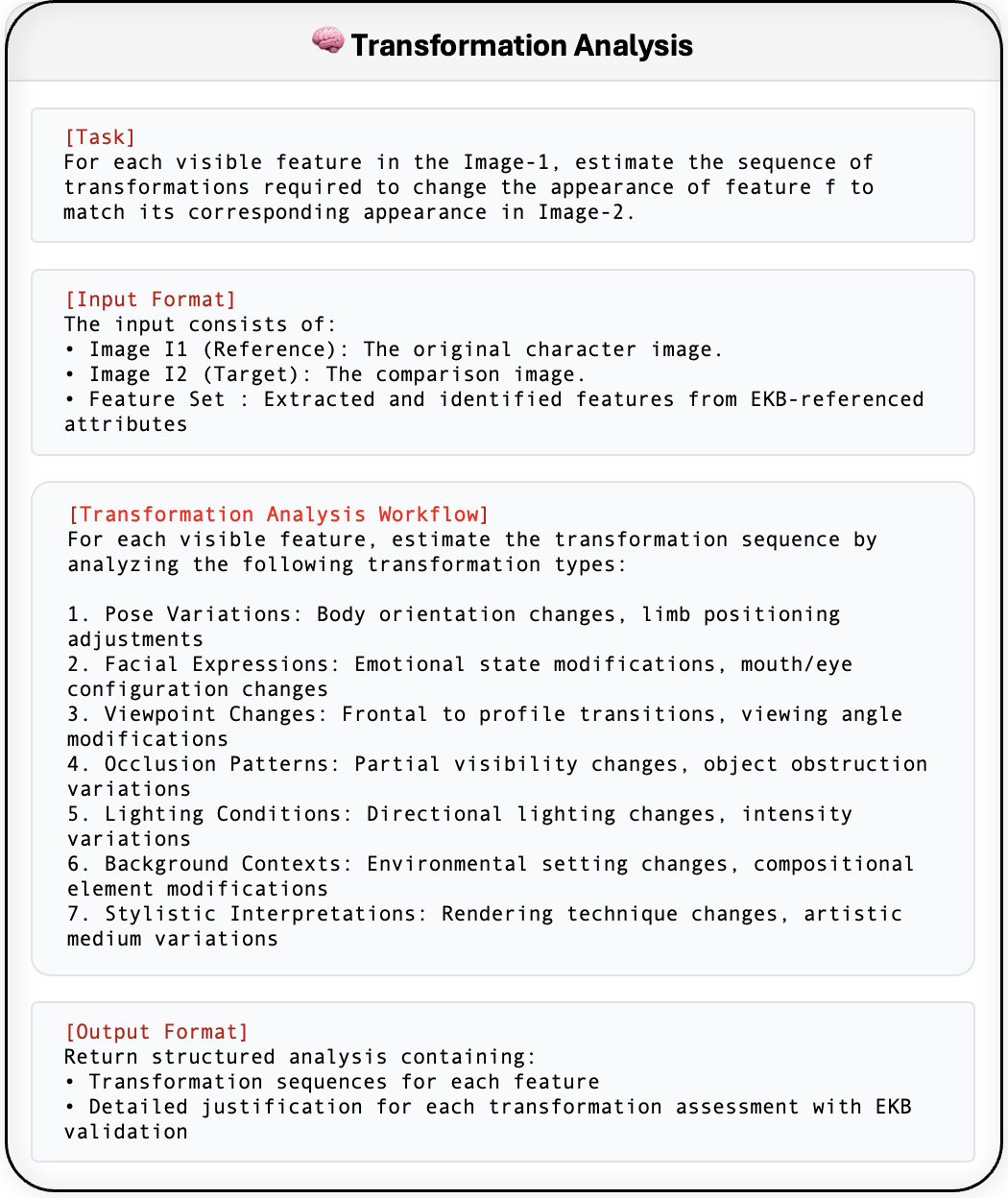}
    \caption{Transformation Analysis Prompt}
    \label{fig:transformation_prompt}
\end{figure}

\noindent
\textbf{Transformation Analysis:} For each visible feature $f \in F_{s_1}$ in the reference image, and the set of transformations $X$, we estimate the sequence of transformations required to change the appearance of feature $f$ from image $I_1$ to match its appearance in image $I_2$. This is accomplished by prompting the VLM: $\rho_f := \theta(I, p)$, where the prompt $p$ is constructed as demonstrated in Figure~\ref{fig:transformation_prompt}, and $\rho_f$ contains both the transformation sequence.

\begin{figure}
    \centering

    \includegraphics[width=0.9\linewidth]{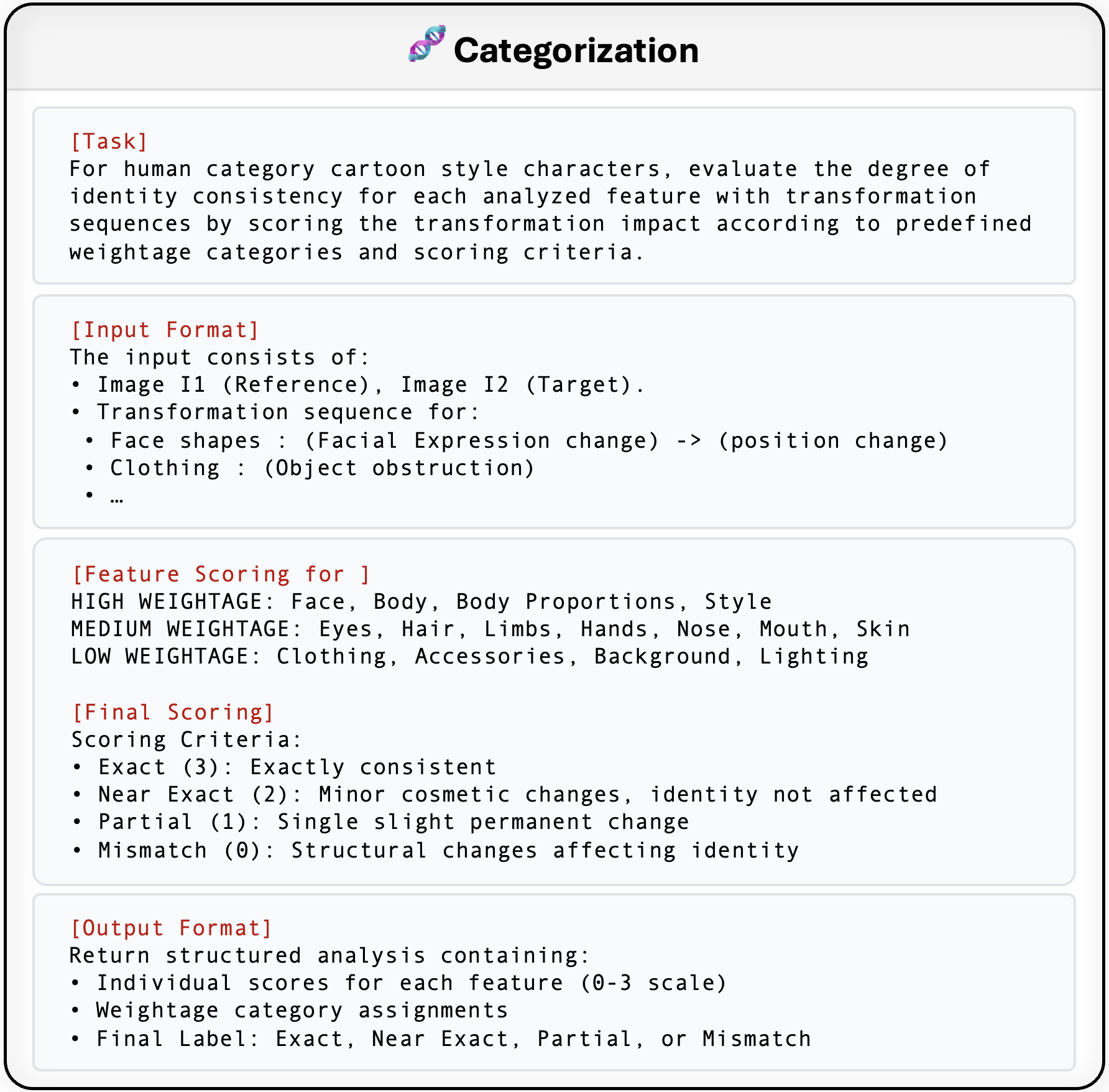}
    \caption{Structured VLM prompt for identity preservation assessment. The prompt guides categorization of image pairs into four consistency classes. Transformation importance rules pre-defined in the context.}
\label{fig:categorization_prompt}
\end{figure}

\noindent
\textbf{Identity Preservation Categorization:} To obtain the final identity preservation assessment, we provide the VLM with domain-specific rules, which encode (i) the relative importance of different features for identity preservation (e.g., facial features weighted more heavily than clothing details) and (ii) which transformations significantly harm identity (e.g., structural changes vs. acceptable style variations). The VLM aggregates individual transformation sequences $\{\rho_f\}_{f \in F_{s_1}}$ using these rules to assign a final identity preservation category:
\begin{equation}
    \varphi(I_1, I_2) \in \mathcal{C}
\end{equation}
This rule-based aggregation ensures consistent categorization aligned with human perception of identity preservation. An example prompt is shown in Figure~\ref{fig:categorization_prompt}.



\section{Benchmark}
Evaluating identity-preserving generation requires benchmarks that capture real-world complexity, a gap that current datasets fail to address. Despite rapid advances in generative models, evaluation remains limited to simple, isolated transformations. We analyze existing benchmark limitations and present a comprehensive dataset that addresses critical gaps through balanced representation of underrepresented subjects and styles, while incorporating compound multi-axis transformations that reveal model failure modes.



\paragraph{Limitations of Existing Benchmarks.}
DreamBench~\cite{ruiz2023dreambooth} and CustomConcept101~\cite{kumari2023customdiffusion} probe identity preservation only along one to two simple transformation axes such as background change, style transfer, accessory change, or single-property edits. DreamBench++~\cite{peng2024dreambenchpp} enlarges the benchmark with harder prompts that compose three-to-four edits at once (e.g. style + scene + time-of-day + new objects). However,
complex compound simultaneous changes to pose, expression, occlusion, and viewpoint remain untested despite representing common failure modes in deployed systems. These multi-axis variations mirror real creative workflows where artists require consistent character identity across diverse scenes and emotional states.

\paragraph{Representational Gaps in Current Data.}
Current benchmarks exhibit severe category imbalances that limit comprehensive evaluation;
\begin{itemize}
    \item \textbf{Style bias:} Photorealistic images dominate existing datasets, while vector art and cartoon rendering, essential for animation and design applications—remain significantly under-represented
    \item \textbf{Subject limitations:} Anthropomorphic characters and inanimate animated subjects receive minimal coverage despite their prevalence in creative industries
    \item \textbf{Transformation sparsity:} Combinations involving extreme viewpoint shifts, heavy occlusions, or dramatic expression changes are rarely tested
\end{itemize}
These omissions create blind spots in model evaluation, preventing accurate assessment of generalization capabilities across domains crucial for practical deployment.
\begin{figure}
    \centering
    \includegraphics[width=\linewidth]{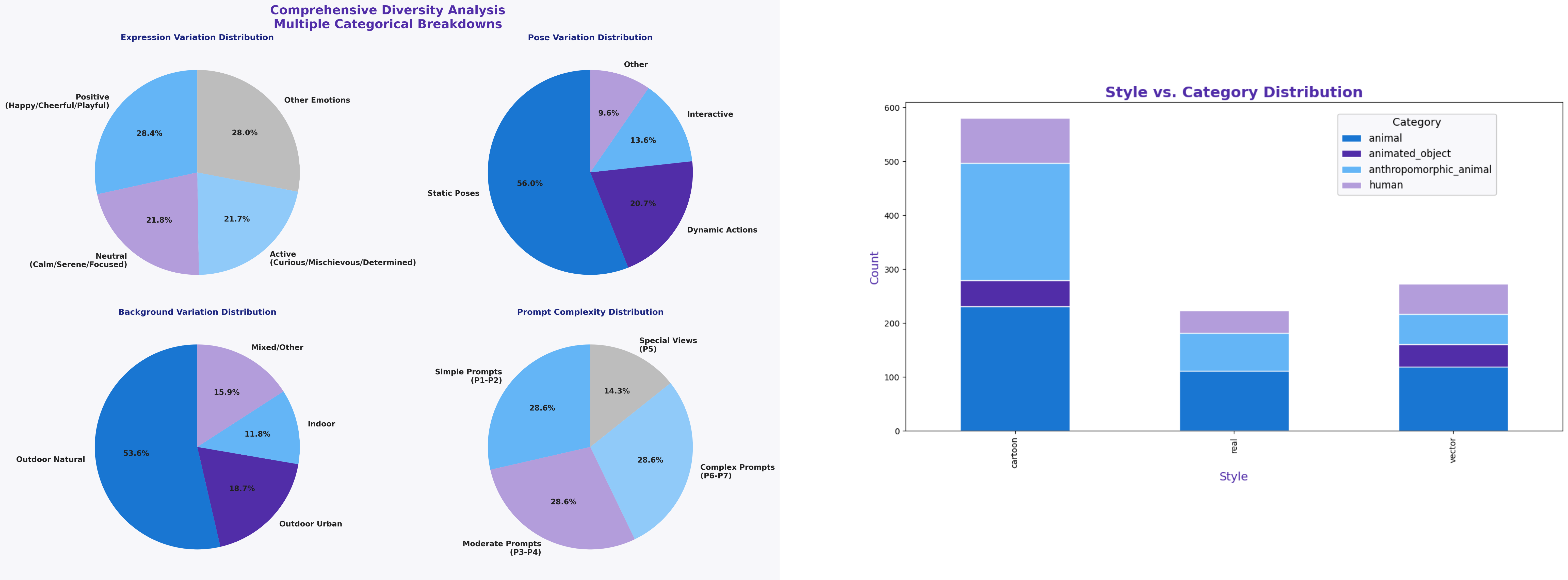}
    \caption{Dataset Diversity across expression, pose, background, and prompt complexity (left). Style-category distribution spanning animal, animated object, anthropomorphic animal, and human(right)}
    \label{fig:benchmark_comp_coverage}
\end{figure}

\paragraph{Comprehensive Benchmark.}
We introduce a new benchmark addressing these systematic gaps through 1,078 carefully designed prompts spanning 154 distinct subjects, each evaluated across seven complex transformation axis, four styles, and four subject types. Our design principles prioritize:
\begin{itemize}
    \item \textbf{High transformation complexity:} Each prompt incorporates 5-6 distinct transformations ($\subseteq X$) applied simultaneously.
    \item \textbf{Mitigating data bias} We place particular emphasis on  anthropomorphic characters and inanimate animated subjects in vector art and cartoon rendering style. 
\end{itemize}
%
%
%
%
Figure~\ref{fig:benchmark_comp_coverage} illustrates the comprehensive coverage of our benchmark relative to prior work.
\begin{figure}
    \centering
    \includegraphics[width=\linewidth]{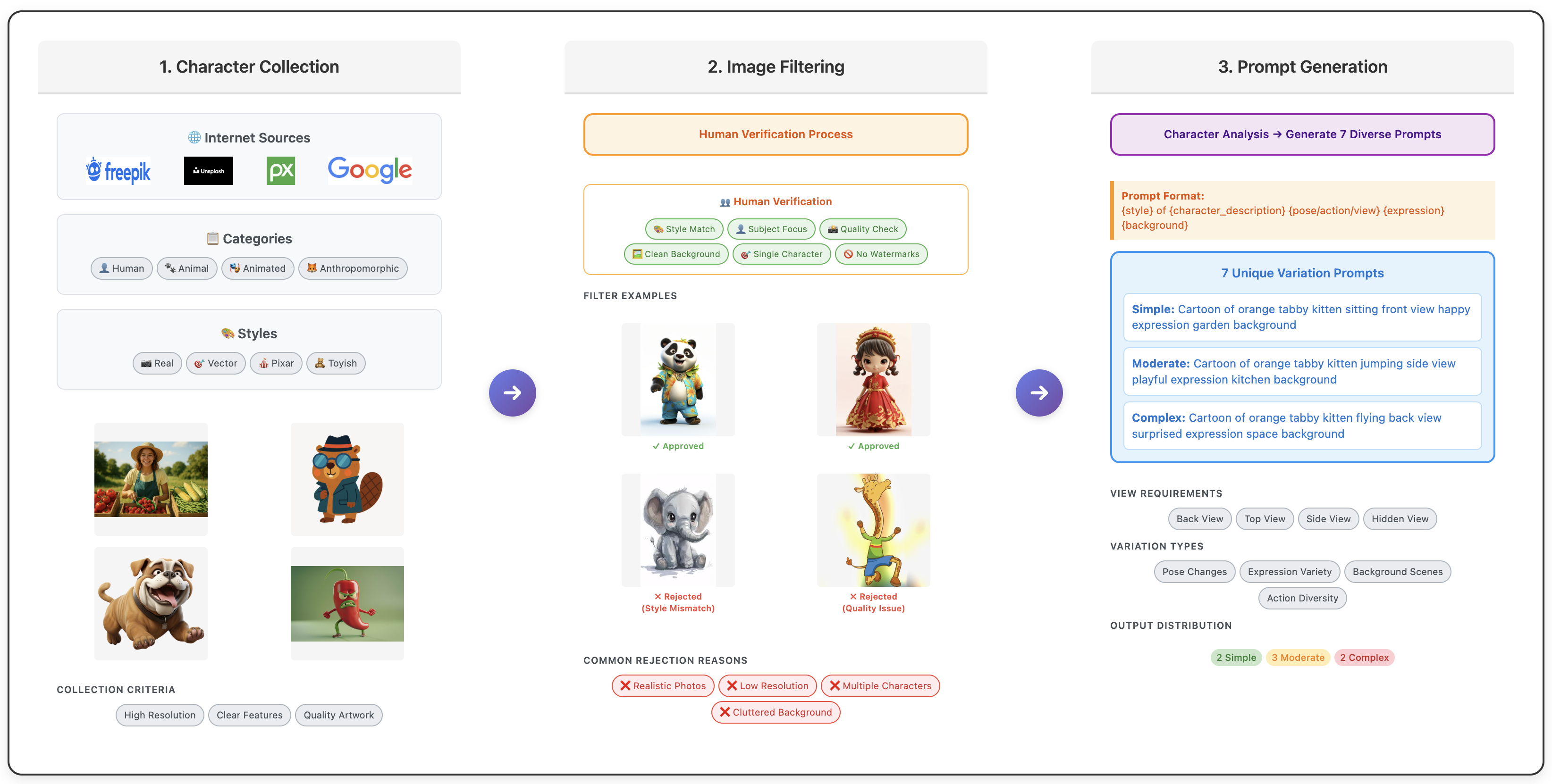}
    \caption{\textbf{Our dataset construction pipeline.} (1) character collection from diverse sources and styles, (2) human-verified filtering based on quality and content criteria, and (3) structured prompt generation with controlled variation in style, pose, expression, and background.}
    \label{fig:bench_mark_creation}
\end{figure}

\subsection{Dataset Construction Methodology}
Our benchmark construction employs a systematic two-stage pipeline that ensures representational diversity and transformation complexity while maintaining quality control through human oversight.

\begin{table}[tb]
\centering
\small
\begin{tabular}{@{}lcccc@{}}
\toprule
\textbf{Model} & \textbf{H--H} & \textbf{G--H} & \textbf{C--H} & \textbf{D--H} \\
               &               & \textbf{(Ours)} & (CLIP) & (DINOv2) \\
\midrule
Anystory & 0.820 & 0.387 & 0.056 & 0.109 \\
DSD      & 0.602 & 0.484 & 0.168 & 0.209 \\
Omnigen  & 0.664 & 0.381 & 0.150 & 0.094 \\
UNO      & 0.655 & 0.419 & 0.189 & 0.246 \\
\bottomrule
\end{tabular}
\caption{Pearson correlations ($r$) between human--human (\textbf{H--H}), our method--human (\textbf{G--H}), CLIP--human (\textbf{C--H}), and DINOv2--human (\textbf{D--H}) aggregated by generative model.}
\label{tab:model_correlations}
\end{table}

\begin{table}[tb]
\centering
\small
\begin{tabular}{@{}llcccc@{}}
\toprule
\textbf{Category} & \textbf{Style} & \textbf{H--H} & \textbf{G--H} & \textbf{C--H} & \textbf{D--H} \\
                  &                &               & \textbf{(Ours)} & (CLIP) & (DINOv2) \\
\midrule
Animal            & Cartoon & 0.651 & 0.472 & 0.069 & 0.138 \\
Animal            & Real    & 0.642 & 0.419 & 0.170 & 0.241 \\
Animal            & Vector  & 0.660 & 0.384 & 0.169 & 0.238 \\
Anim.\ Obj.       & Cartoon & 0.687 & 0.437 & 0.137 & 0.180 \\
Anim.\ Obj.       & Vector  & 0.717 & 0.372 & 0.168 & $-0.071$ \\
Anthro.\ Anim.    & Cartoon & 0.672 & 0.391 & 0.163 & 0.150 \\
Anthro.\ Anim.    & Real    & 0.827 & 0.537 & 0.168 & 0.262 \\
Anthro.\ Anim.    & Vector  & 0.727 & 0.533 & 0.146 & 0.140 \\
Human             & Cartoon & 0.829 & 0.503 & 0.200 & 0.340 \\
Human             & Real    & 0.757 & 0.487 & 0.377 & 0.314 \\
Human             & Vector  & 0.736 & 0.434 & 0.125 & 0.106 \\
\bottomrule
\end{tabular}
\vspace{0.3em}
\footnotesize \textit{Anim.\ Obj.~$=$ Animated Object,\; Anthro.\ Anim.~$=$ Anthropomorphic Animal.}
\caption{Pearson correlations ($r$) between human--human (\textbf{H--H}), our method--human (\textbf{G--H}), CLIP--human (\textbf{C--H}), and DINOv2--human (\textbf{D--H}) across category–style pairs.}
\label{tab:cat_style_correlations}
\end{table}



\begin{table}[tb]
\centering
\small
\begin{tabular}{@{}lccccc@{}}
\toprule
\textbf{Model} & \textbf{$\bar{h}$} & \textbf{$\bar{g}$} & \textbf{$\bar{c}$} & \textbf{$\bar{d}$} \\
               & (Human)           & (Ours)             & (CLIP)             & (DINOv2)           \\
\midrule
Anystory & 0.629 & 0.588 & 0.830 & 0.691 \\
DSD      & 0.330 & 0.403 & 0.804 & 0.632 \\
Omnigen  & 0.201 & 0.341 & 0.830 & 0.664 \\
UNO      & 0.579 & 0.569 & 0.834 & 0.703 \\
\bottomrule
\end{tabular}
\caption{Average \emph{normalized} scores per model.  
$\bar{h}$ is the mean of the two normalized human ratings;  
$\bar{g}$, $\bar{c}$, and $\bar{d}$ are the corresponding averages for our method, CLIP, and DINOv2, respectively.}
\label{tab:model_avg_norm}
\end{table}

\paragraph{Stage 1: Reference Image Acquisition.}
We sampled across subject-style combinations to ensure balanced representation manually. High-resolution reference images were sourced from royalty-free repositories (Freepik, Unsplash) and filtered according to strict criteria:
\begin{itemize}
    \item \textbf{Subject isolation:} Single character focus with minimal background distractions
    \item \textbf{Quality requirements:} Resolution $\geq$ 1024×1024, absence of compression artifacts or watermarks
    \item \textbf{Pose diversity:} Varied initial poses to enable meaningful transformation testing
    \item \textbf{Style authenticity:} Clear adherence to target artistic style without ambiguity
\end{itemize}
See Figure~\ref{fig:bench_mark_creation} for an example of the creation process.
The resulting dataset comprises equal representation across 12 combinations of categories (4 subject types × 3 artistic styles), addressing the systematic biases present in existing benchmarks. 

\paragraph{Stage 2: Transformation-Rich Prompt Synthesis.}
Prompt generation follows a hybrid automated-manual pipeline. \newline
\textbf{Automated Initial Generation:} GPT-4o analyzes each reference image to extract attributes and generate plausible transformation scenarios (the mechanism is similar to hierarchical decomposition). 
The model is instructed to generate prompts that incorporate 5-6 simultaneous transformation axes per prompt. For example, prompts with complex pose specifications involving coordinated multi-limb movements, viewpoint variations spanning extreme angles (back view, top-down, worm's-eye), expression changes and environmental interactions.\newline
\textbf{Expert Refinement Protocol:} Human reviewers iteratively refine the generated prompts to ensure that each prompt incorporates complex, compound transformations that are physically plausible and suitably challenging. Reviews also filter out redundancy across different transformation, type and style.
%
The resulting prompts significantly exceed prior benchmarks in complexity, with quantitative analysis confirming an average of 5.4 transformation axes per prompt compared to 2.1 in DreamBench and 3.2 in DreamBench++.

\section{Experiments}

\paragraph{Setup:}
We evaluate CHARIS on four state-of-the-art single-image character generation models: \emph{UNO}~\cite{wu2025less}, \emph{DSD}~\cite{cai2024diffusion} (Flux.1-dev based), \emph{AnyStory}~\cite{he2025anystory} (SDXL based), and \emph{Omnigen}~\cite{shitao2024omnigen} (Phi-3 based). All models require only one reference image and a text prompt, without inference-time fine-tuning. We use official implementations with default parameters.
\emph{Eight expert annotators} (two per model) assessed generated images using four consistency categories:
\begin{itemize}
\item \textit{Exact Match}: Full identity preservation
\item \textit{Near Exact Match}: Minor cosmetic variations not affecting identity
\item \textit{Partial Match}: Significant alterations but retaining identifiable features
\item \textit{Mismatch}: Severely compromised or lost identity
\end{itemize}
Annotators were provided with the same structured guidelines as the VLMs. Inter-annotator agreement, measured via Pearson correlation, ranged from approximately $0.6$ to $0.8$, establishing an empirical upper bound for comparison with VLM performance (Table~\ref{tab:cat_style_correlations}).

\begin{figure}
    \centering
    \includegraphics[width=\linewidth]{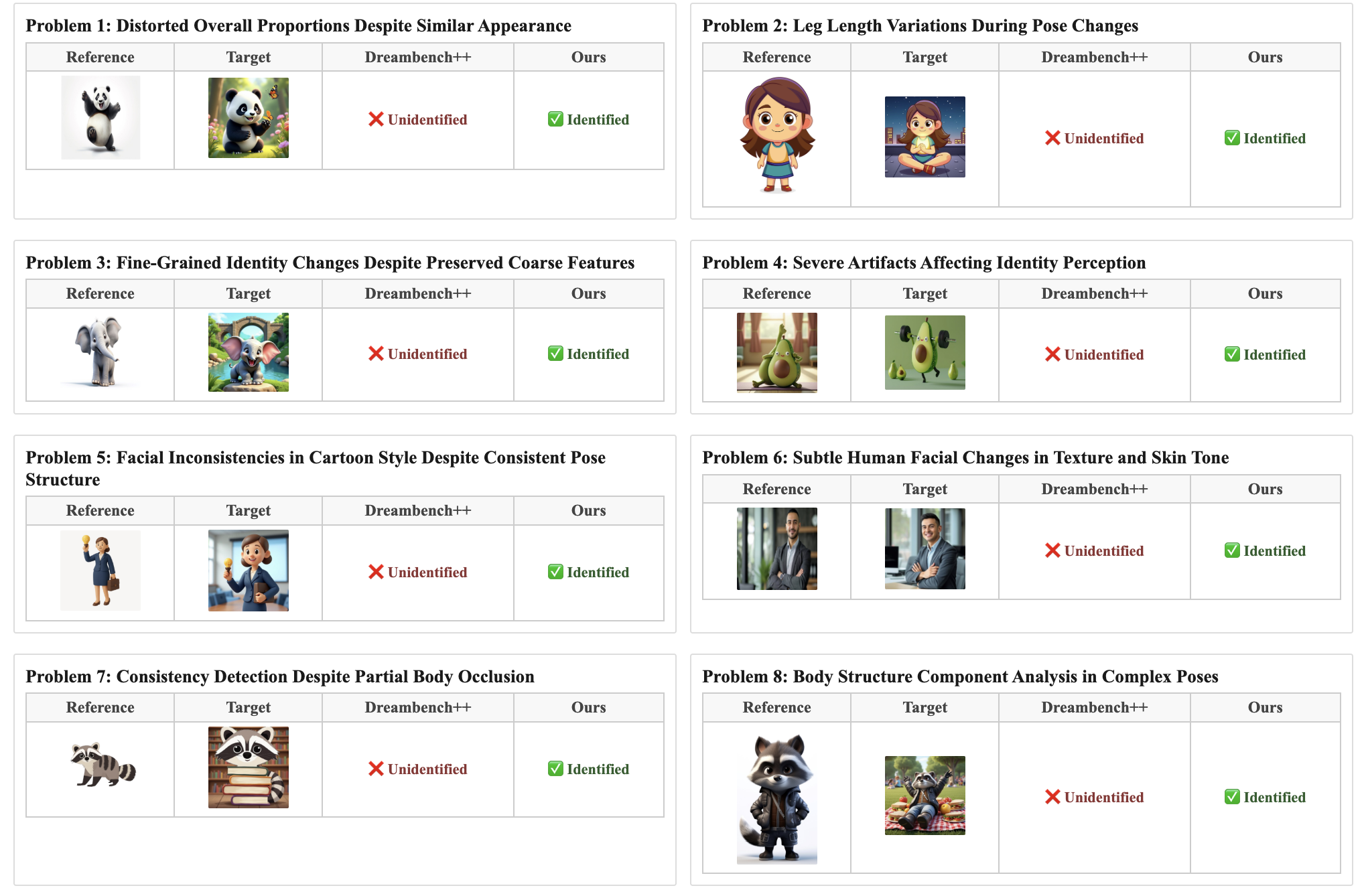} 
    \caption{Image Consistency Detection Performance Comparison. Our method successfully identifies consistency issues (Identified) across different challenging scenarios where Dreambench++ fails to detect subtle inconsistencies (Unidentified)  between reference and target image pairs.}
    \label{fig:qualitative_insights}
\end{figure}

\paragraph{Qualitative Insights:}
Our qualitative analysis demonstrates CHARIS's effectiveness across diverse evaluation scenarios, (see Figure~\ref{fig:qualitative_insights})
\begin{itemize}
    \item \textit{Fine-Grained Facial Analysis:} More accurate identity assessment on small or stylized (cartoon) faces via attribute-level decomposition.

    \item \textit{Body Proportion Detection:} Better at identifying proportion-based inconsistencies across visual styles, and remain robust under substantial pose variations.

    \item \textit{Artifact vs. Identity Drift:} 
    Disentangles generation artifacts from genuine identity drift, reducing common misclassifications.
    

    \item \textit{Occlusion Handling:} Separates occlusion-induced discrepancies from other attribute changes.

    \item \textit{Expression Invariance:} Does not penalize expression variation.

    \item \textit{Style Sensitivity:} 
    Flags subtle shifts in artistic style, rendering, and texture despite high global similarity.
\end{itemize}

\paragraph{Correlation Study.}
To assess alignment with human perception, we compute Pearson correlations between human ratings (H-H), our method (G-H), and baselines—CLIP (C-H) and DINOv2 (D-H)—across subject categories $\mathfrak{T}$ and styles $\mathfrak{K}$. Table~\ref{tab:cat_style_correlations} shows that G-H consistently outperforms the baselines, particularly for \emph{anthropomorphic animals} and \emph{animated objects} in \emph{vector/cartoon} styles. In the \emph{human-cartoon} evaluation, our method (G-H = 0.50) approaches human agreement (H-H = 0.83), while CLIP and DINOv2 lag significantly (C-H = 0.20, D-H = 0.34). Embedding-based metrics often fail to capture identity-preserving variations such as proportion shifts or design reinterpretations, with the largest failures observed in animated object and animal-cartoon scenarios where structural changes are common.
Across both \emph{correlation} and \emph{average-score} metrics (Tables~\ref{tab:model_correlations} and~\ref{tab:model_avg_norm}), \textsc{Uno} demonstrates the most consistent and balanced performance, showing strong alignment with human ratings and high identity preservation.
\textsc{Anystory} achieves the highest consistency score in our benchmark (Table~\ref{tab:model_avg_norm}). Nevertheless, its outputs exhibit subtle identity shifts that human raters detect but our VLM-based method often overlook (Table~\ref{tab:model_correlations}).
\textsc{Omnigen} performs well on CLIP and DINOv2 but receives the lowest human ratings, suggesting that it often preserves visual similarity without maintaining perceived identity.

\section{Conclusion}
We presented CHARIS, a hierarchical evaluation framework that addresses fundamental limitations in identity preservation assessment. By decomposing evaluation into feature-level transformations, our approach grounds VLM analysis in verifiable visual evidence, improving consistency evaluation. Our benchmark of 1,078 prompts fills critical gaps with balanced coverage of underrepresented categories (anthropomorphic, animated) and styles (vector, cartoon), incorporating 5-6 transformation axes per prompt.

Empirical results demonstrate superior correlation with human judgments compared to embedding-based metrics, particularly for stylized content where traditional approaches fail. The framework's interpretable transformation categorization provides actionable insights for model development, while our analysis of four state-of-the-art models reveals distinct capabilities and failure modes. 
\paragraph{Limitations.}
While our VLM-based evaluation demonstrates significantly improved alignment with human perceptual judgments compared to existing baselines, certain limitations remain. Specifically, our approach still struggles to reliably detect subtle identity-preserving details—such as minor proportional adjustments, slight stylistic variations, or fine-grained object features—that humans readily perceive but which exceed the inherent resolution capabilities and semantic abstraction limits of current vision–language models. Additionally, nuanced human preferences and context-dependent interpretations of identity consistency can be challenging for VLM-based assessments, underscoring limitations in capturing subjective human judgment. These challenges highlight important avenues for future research aimed at refining model granularity and incorporating richer contextual understanding into identity consistency evaluations.


\paragraph{Future Work.}
Our framework currently evaluates only single-subject identity preservation, whereas real-world applications increasingly require multi-subject consistency, which is a significantly more complex challenge. Extending CHARIS to multi-subject scenarios requires addressing identity entanglement, role confusion, and interaction coherence among subjects. This extension necessitates expanding our \ekb~to include relational attributes (e.g., spatial arrangements and interactions) and cross-subject transformation rules (e.g., synchronized poses and complementary expressions).

\bibliography{aaai2026}

\appendix

\end{document}